\documentclass[sigconf]{acmart}

\AtBeginDocument{%
  \providecommand\BibTeX{{%
    \normalfont B\kern-0.5em{\scshape i\kern-0.25em b}\kern-0.8em\TeX}}}

\renewcommand\footnotetextcopyrightpermission[1]{} 
\makeatletter
\let\@authorsaddresses\@empty
\makeatother
\begin{document}

\title{Gender and Robots: A Literature Review}

\author{David Gray Widder}
\email{dwidder@cmu.edu}
\affiliation{%
  \institution{Carnegie Mellon University}
}

\renewcommand{\shortauthors}{Widder}

\begin{abstract}
  Here, I ask what we can learn about how gender affects how people engage with robots. 
  I review 46 empirical studies of social robots, published 2018 or earlier, which report on the gender of their participants or the perceived or intended gender of the robot, or both, and perform some analysis with respect to either participant or robot gender. 
  From these studies, I find that robots are by default perceived as male, that robots absorb human gender stereotypes, and that men tend to engage with robots more than women. I highlight open questions about how such gender effects may be different in younger participants, and whether one should seek to match the gender of the robot to the gender of the participant to ensure positive interaction outcomes. 
  
  I conclude by suggesting that future research should: include gender diverse participant pools, include non-binary participants, rely on self-identification for discerning gender rather than researcher perception, control for known covariates of gender, test for different study outcomes with respect to gender, and test whether the robot used was perceived as gendered by participants. I include an appendix with a narrative summary of gender-relevant findings from each of the 46 papers.
\end{abstract}


\maketitle

\section{Motivation and Method}
In service of a separate study, Wang et al. found in 2014 that 
of 190 papers published in the Conference on Human Robot Interaction 
only 106 (56\%) provided the sex breakdown of their participants, 
and within those, only
21 (20\%, 11\% overall) ``provided at least minimal or passing sex-based 
analysis''~\cite{wang2014beyond}.
However, we know that gender has important effects in human-human 
interaction, so as robots learn to interact with and emulate 
humans better, it stands to reason that gender will increasingly affect these interactions.
Despite this importance, I have not found a literature review
of empirical studies on gender in robotics. 

In this paper, I attempt to establish what ``big facts''
are known, only concluding when there appears to be a significant
body of work in agreement. 
I limit my scope to social robots. 
Notably given the topic of gender, this excludes studies on sex robots, 
which has seen significant theoretical and 
empirical treatment elsewhere.
I limit my scope to empirical pieces involving human participants, thus excluding critical or theoretical pieces,
though I have read those to find more empirical studies and to inform my thinking, 
and robot design papers which have not been evaluated by participants 
in some way.
I note that these studies tend to 
be quantitative experimental studies (with one exception~\cite{siino2005robots}).
I found the included studies by using Google Scholar
to find studies which included the words ``gender'' 
and ``robot'' in the title or abstract,
and retaining both studies which modify the robots gender
or analyze with respect to the gender of their human participants. 
This review was conducted in 2018, and there have undoubtedly been many gender relevant robotics papers published since. 
This review not conducted using a Systematic Literature Review (\textit{ie}, ~\cite{calvaresi2018multi}), so it may not be exhaustive. 
Nonetheless, I hope that this review of 46 studies on gender in robotics may nonetheless prove useful as a starting point for further reading. 

I try to maintain a critical tone when reporting conclusions drawn from 
the papers reported here, especially as a result of concerning statistical choices I saw  
in many papers, such as biviariate correlations with no controls for confounds, 
reporting non-significant results, 
or not reporting effect sizes.

Almost all of the work I read assumed a male-female gender
binary, which is not reflective of reality.
However, because of this, my literature review
cannot comment on the experiences of non-binary people,
and thus my work is restricted to this binary. 

The headings of the next three sections reflect the
three broad conclusions I feel comfortable drawing from these papers:
1) by default, robots are perceived as male, 2)
robots absorb human gender stereotypes, 
3) men tend to engage with robots more than women. 

I conclude with a practical set of guidelines for robotics researchers, 
and in the appendix I leave a narrative summary of each paper I read, 
which may be of interest, but is very verbose.

\section{By default, robots perceived as male}
By default, I find that people will think robots are male, even in
the absence of intentional gender cues, or even 
just when thinking about robots in the abstract.
One study asked older adults to draw a robot, most tended to draw
a robot with male or gender neutral features~\cite{stafford2014older}.
Another study asked robots to name a robot, 
and found that a mixed gender participant pool assigned the 
robot mainly neutral or male names, with only 1\% assigning it a female name~\cite{walters2008human},
even though this robot had no anthropomorphic features, and thus no explicitly gendered cues.
An in depth ethnographic study of a robot deployed in a hospital found 
that higher status males tended to be excited about the robot, 
but see it as a non gendered machine,
whereas women and lower status males tended to see it 
as something out of their control, 
not likely to help them perform their jobs (one of its designed intentions)
and for the entertainment of the higher status men who
acquired it~\cite{siino2005robots}, and they referred to it using male pronouns despite
it being non-anthropomorphic and lacking explicit gender cues.

\section{Robots absorb human gender stereotypes}
Robots can be endowed with their creators with a gender, 
with consistent effects on the interactions they have with participants.
For example, one study found that male participants were more likely to donate to a robot
with female gender cues~\cite{siegel2009persuasive}.
However and worryingly, human stereotypes often appear to transfer to gendered robots:
one study shows that a large sample of 84 men and 79 women preferred 
a healthcare robot when it was female presenting, and a security
robot when it was male presenting, aligning with the ``female-nurturing''
and ``male-controntational'' stereotype~\cite{tay2014stereotypes}.
Similarly, another study found that having a robot's voice 
to align with the gender stereotype between ``male'' functional tasks
and ``female'' social tasks made this communication more 
effective~\cite{rhim2014effect}.

Aligning with psychological theory the author cite that women are perceived as warm whereas men are perceived as having agency, one study found that men had an ``uncanny valley''
reaction with a female-cued robot which aligned with her gender stereotype,
but women have an similarly negative reaction to a male cued robot aligned with his 
stereotype, suggesting that aligning with the gender stereotype is a way
to heighten anthropomorphism~\cite{otterbacher2017s}.

One study found that when being instructed on a typically female versus male task by a robotic collaborator, 
people made more errors, were less willing to accept help from the robot on a future task, 
and anthropomorphize the robot to a lesser extent.
This held regardless of the gender cues of the robot (even when
a female cued robot instructed on female stereotyped tasks), 
suggesting that they viewed ``male'' as the natural gender of the robot, 
and that robots are better suited to work with people to complete male stereotyped tasks~\cite{kuchenbrandt2014keep}.
However, another study appears to suggest that a mismatch between 
the gender of the robot and the gender stereotype associate with a task 
may increase increased willingness 
to be taught by an instructor robot during learning tasks~\cite{reich2017ir}.

One study found that Roombas are more likely to be given
as gifts to women than men, using this to propose the previous 
effect: vacuum's association with ``women's work'' housework
shifts their gender association from the default male (above) towards the feminine, despite
lacking anthropomorphic features or explicit gender cues, 
though roombas were as likely to be given female names as male names~\cite{sung2008housewives}.

In an example of appearance-based gender norms, one study found that men and women discussed a highly
anthropomorphic robot's female gender, and female stereotyped qualities such as ``pretty'' and ``comforting''
when explaining why they would feel more comfortable letting it in their home 
than a more mechanical robot~\cite{carpenter2009gender}, 
while one related study did not find a statistical robot gender when asking 
people if they'd let a robot in their home, but the authors acknowledge
this may be due to having few participants and thus low statistical power~\cite{dautenhahn2005robot}.
Another study found that people speak quieter to a female robot, 
even when it spoke louder~\cite{strupka2016influence}.

A study with children found that they also recognize the 
gender of robots, and apply the same norm of ``gender-segregated''
play with robots as they do with other kids~\cite{sandygulova2016investigating}.

In an example of showing that ability  based gender norms transfer to robots,
two researchers gendered a robot using cues such as longer hair for
female or a hat for male, 
they found that participants found math tasks more suitable
for the male cued robot, and verbal tasks more suitable for the female robot,
and that they found the male robot to have more agency, 
whereas the female robot was more communal~\cite{eyssel2012s}.
Another such example found that participants assumed a female cued robot
knew more about dating norms, a topic they believe is stereotypically female, 
leading participants to use more words when describing these norms to male cued
robots.~\cite{powers2005eliciting}.

As shown, robots can be used to perpetuate gender norms 
and harmful gender stereotypes. 
Even though gender norms can be used to improve human robot interaction, roboticists should resist exploiting this advantage, 
in order to improve society,
even if aligning with, and thus perpetuating gender norms, 
helps their robots interact better in the short term.

\section{Men tend to engage robots more than women}
People of different genders react to robots differently. 
Men tend to like, be more comfortable with, and engage with robots more than women.
For example, in an evaluation of a social robot for eldercare, 
one study found that men rated the robot higher 
in all conditions higher than women with a medium 
effect size~\cite{stafford2014older}.
The affect of men engaging with robots more 
appears to overpower gender norms in some cases:
men were more likely to ask for help from a robot than women,
despite the assumption that men are less likely to ask 
for help~\cite{alexander2014asking}.
Another study found that men like robots more than women, 
and women identify with feelings of robophobia more
than men~\cite{halpern2012unveiling}.
Another study found that men feel more positively
about enganging robots in a healthcare setting 
than women~\cite{kuo2009age}.
For example, having a robot's head facing towards men makes
them more comfortable when it approaches them, but makes women less comfortable~\cite{takayama2009influences}, 
and another study found even more gendered differences in the 
preferred method of robot approach~\cite{dautenhahn2006may},
and another found that men were more comfortable with the robot
approaching closer when it did so from the side than the front,
but that women let it approach from the front to a closer distance
than men did~\cite{syrdal2007personalized}.
Another study placed a robot in a public place, and found
that men approached the robot closer than the women did~\cite{van2008visual}. 
Men answer surveys in more socially desirable ways
and also perform worse on tests when
they are administered with a robot
rather than on paper
suggesting that men see robots as more of a
social, human like presence than women~\cite{schermerhorn2008robot}, 
though another study found that men 
perform in less socially desirable ways
when interacting with a robot rather than a
disembodied voice~\cite{crowelly2009gendered}.
The ethnographic study referenced above found
that males were excited about what a robot could do,
and appeared to see it as entertaining
whereas females saw it as distracting,
and not helpful for their work~\cite{siino2005robots}.
Another study found that when interacting with 
robots in diads, conversation time was most imbalanced
with older males and younger females, with males dominating
interaction with the robot~\cite{skantze2017predicting}.
Offering one possible explanation for the mechanism why
men tend to engage with robots more, 
one study found that men in their sample 
were more likely to have experience using 
computers, and also a higher perceived ease of 
use with a healthcare robot~\cite{heerink2011exploring}.
Another study found that men were more willing to be aggressive
when instructed to kill a robot than women were~\cite{bartneck2007kill}.
One study found that women rate a robot worse
when it looks at them more, whereas men did the 
reverse~\cite{mutlu2006storytelling},
and another found that women were more apprehensive
about approaching robots which appeared to have broken 
down~\cite{rahimi1990human}.
One study found that fathers were more positive about
using robots in their kids education than mothers, 
and more willing to
help their kids use them~\cite{lin2012exploring}, 
and another found that when working on robotics
projects in groups, girls were more likely to 
chose to program, blog about, or build websites
for the robot, leaving boys to build the robot~\cite{weinberg2007impact}.
Men and women sometimes believe that a robot
mirrors their personality, but that the kinds of 
personalities they believe the robot mirrors is different
based on the participant's gender~\cite{woods2005robot}.

Perhaps offering another explanation for how to 
engage women with robots, one study found that 
women rate robots higher when they are polite~\cite{strait2015gender}, 
but also, contradicting other results in this section, 
expressed a greater interest in interacting with the 
robot than men.
Again contradicting past results, another
study appeared to find that women were more likely to stand
closer to a robot, and had more positive attitudes
towards the idea of robots having emotions~\cite{nomura2006experimental}.

Despite some conflicting evidence, it appears
that men are more willing to engage with robots.
After reading these studies, the most convincing reasons for
this disparity I found were 
robot \textit{self efficacy}, and level of \textit{past experience}, 
two constructs which often explain gender norms in other contexts.
I suggest that future work investigate this by 
controlling for these two factors, to see if a gender effect remains.

\section{Open question: But what about the children?}
Gender norms are socialized as one grows up. 
Future work can use robots as a tool to interrogate the formation of these gender norms.
Girls seem to feel more social and attraction to 
anthropomorphic robots, but boys tend to be more attracted 
to machine like robots~\cite{tung2011influence}.
Further, one study found that boys tend to act more aggressively to a 
cat like robot, whereas girls are more soothing~\cite{scheeff2002experiences}.
When interacting with robots in diads, conversation time was most imbalanced
with older males and younger females, with males dominating
interaction with the robot~\cite{skantze2017predicting}.
One study found that when interacting with a robot in diads, 
younger participants allowed the older participant to 
interact with the robot more~\cite{skantze2017predicting}
One study suggested that boys were more expressive when 
interacting with a robot than girls, because it was
easier to classify based on their reaction whether they had won
or lost a game against a robot~\cite{shahid2010child}.

Given that gender norms are sometimes different for kids, and that many adult 
gender norms are socialized during childhood, 
I suggest that future HRI work can further investigate age-gender interaction
effects.

\section{Open question: should one match robot-participant gender?}
Sometimes it is better to match the gender of the participant to the gender of 
the robot. Sometimes, the reverse is better. Choosing between the two is hard, 
and appears to depend on the particularities of the situation, gender norms.
Four studies surface this:
In one study, males were faster when performing the task than females, while this was only true when they interacted with a same-gender robot~\cite{kuchenbrandt2014keep}. 
Another found that people generally rated
robots of their opposite gender as more credible~\cite{siegel2009persuasive}, 
whereas another found weak evidence that the 
reverse was true: men like male robots more,
and women like female robots more~\cite{eyssel2012s}.
One study found that younger children prefer a robot
of a matching gender, with no effect found 
for older children or adults~\cite{sandygulova2018age}.
Another study found that when working with a robot to complete
a Sodoku task, participants prefer 
working with robots of the opposite gender~\cite{alexander2014asking}.
Future work should figure out better guidance or a 
theoretical model for when to match robot-participant gender,
and when to not. 
Future work should further investigate the circumstances in 
which matched or unmatched robot-human genders work best.








\section{Reflecting on the studies: My suggestions}
Based on my own experience conducting research on gender
and technology, 
and after reflecting on statistical concerns I saw when conducting this review, I propose these guidelines:

Suggestions on Gender: 

\begin{enumerate}
\item Recruit gender diverse participant pools,
\item Explicitly try to recruit non-binary participants,
\item Ask people how they identify, don’t assume or measure gender based on the perception of the researcher,
\item Measure \& control for known correlates of gender (computer self efficacy, past experience with tech \& robots),
\item Analyze outcomes with respect to participant gender,
\item Try to make sure your research team is gender diverse (especially if studying gender!),
\item Test and report whether your participants perceived your robot as gendered, even if you did not attempt to explicitly gender it, to aid future work on gender in robotics. 
\end{enumerate}

Suggestions on Statistics: 
\begin{enumerate}
\item Don’t report effects that are not significant at the 0.05 level (or if you want to, justify your choice of \textit{p} value, or justify results as significant in some other way)
\item Avoid testing for bivariate correlations, instead consider statistical methods which can control for confounding covariates, such as regression 
\item Measure, report, and discuss effect sizes, so that people can evaluate the real world impact of your results.
\end{enumerate}

\section{Appendix: All the Studies}
Here I report all of the studies I read, and their main results 
with respect to gender. 
This may help serve as a literature index.
They are organized by 1) studies which only consider the gender of the robot, 
2) studies which only consider the gender of human participants,
and 3) studies which do both.

\subsection{Gender Of Robots}
In a CHI extended abstract, Jung et al. test the effect of gender cues (pink earmuffs 
for female, a men's hat for male, or no cues) on the gender of a robot as perceived by 144 undergraduates~\cite{jung2016feminizing}.
They found that people find the robot with female cues is perceived as female, and the robot
with no cues is perceived as male, and that one with male cues is perceived as even more male.
The authors use these results to conclude that robots are by default perceived as male, whereas I believe this better supports the idea that the \textit{single} specific robot they use is perceived as male. 
The authors neglect to report participant's gender, which is puzzling given 
the subject of the study and that they collect other demographics such as age and race.

Orefice et al. studied the effect of robot handshake firmness and movement on
ascribed robot gender as perceived by 11 female and 25 male participants~\cite{orefice2016let}.
They recorded the handshakes of extroverted and introverted male and female participants, 
had a robot emulate these, and found that these secondary participants
could successfully perceive the original gender and level of extroversion of the
handshake originator.
While they include participants of different genders, they do not analyze
with respect to participant's different genders. 
They discuss effect size.

\subsection{Gender Of Participants}

Takayama and Pantofaru studied the impact of different approach situations, notably
a participant approaching a robot, an autonomous robot approaching a participant,
and a teleoperated robot approaching a participant. 
They found that when a robot's head is oriented towards 
the participant, it makes women less comfortable having it near them, 
but makes men more comfortable having it near them.
The authors do not report effect sizes~\cite{takayama2009influences}.

In a series of experiments conducted by Tung to investigate children's reactions 
towards robots with varying degrees of anthropomorphism, 
she found that 
girls felt more social and physical attraction to human-like robots, 
especially female robots, whereas boys felt more attraction to mechanical looking robots~\cite{tung2011influence}.
They do not dicsuss effect size. 

When studying robot smiling behaviors Chung-En had 92 male and 141 
female hotel guests in Macau 
look at a digitally manipulated photo of a robot
or human head with different levels of head 
tilt~\cite{chung2018humanlike}. 
She found that male and female participant's ratings of robots
differed in many respects, finding that younger females rated robots
higher than their male counterparts, but the author recognizes
that the fact that human staff photos were all female, 
introducing a possible confound.
They do not discuss effect sizes. 

Scheeff et al. qualitatively report the experiences of 15 girls and 15 boys
interacting with a wheeled cat like robot which could display emotions 
both in the lab, and even more 
participants in a public science museum.~\cite{scheeff2002experiences}. 
They report that older boys tended to act more agressive or harmful towards
the robot, whereas older girls were gentle, said kind things to it, and 
were soothing. 

Lin et al. surveyed Taiwanese 39 parents about their attitudes towards
the use of educational robots in their 
kid's learning environments~\cite{lin2012exploring}.
They found that men (fathers) were more positive toward using 
educational robots than females (mothers), 
with respect to their usefulness, willingness to help 
their kids with robots, and confidence doing so.

Weinberg et al. studied the impact of school robotics programs on 
girl's self efficacy and future career interests in STEM fields, 
and found that they have a positive impact on 
both~\cite{weinberg2007impact}.
However, they also found that girls were more likely to 
choose to program, blog about, or create presentations about instead of build the robot, because they perceived these tasks as easier.
In mixed gender teams with good mentors, they found that girls were more likely to have increased confidence and expectations of 
success in science and math, but this effect was not observed 
for all girl teams.

Schermerhorn et al. studied the social presence of robots
on 24 men and 23 women while completing short tasks:
a survey on their perceptions of robots, 
a standard measure on social desirability, 
and easy and hard arithmetic tasks~\cite{schermerhorn2008robot}.
They found that when the robot administered a survey 
vocally, men answered in a more socially desirable 
way than when they answered the same survey on paper. 
They further found that men's scores
on the arithmetic tasks were negatively affected by 
the robot's presence, whereas this was not the case for 
females.
They use this to conclude that men see the robot as 
more of a human peer than do females, 
thought they acknowledge that the robot's distinctly 
male voice may have affected this result. 

Siino and Hinds conducted an ethnographic study 
set in a community hospital in Northern California, 
which had just acquired and was attempting, 
at first unsuccessfully, to deploy a robot~\cite{siino2005robots}.. 
They conducted approximately 100 hours of observation
and interviews primarily of a male lead ``robot users group''
responsible for the robot's acquisition, use and configuration. 
The robot was able to make deliveries, and had a touch screen and 
keypad, and could use either a male or female voice to communicate.
The robot did not appear to be anthropomorphic. 
They found that men who held engineering and high ranking 
administration jobs primarily saw the robot as a machine,
using words like ``vehicle'', ``computer'', suggesting
that they saw it as under human control. 
They found that female directors of majority female departments such
as Admitting, Medical Records and Media Relations joined female food
service workers and low-status female pharmacy technicians  
anthropomorphize the robot as a human male, 
even before seeing or interacting with the robot.
The authors speculate that these women saw the robot as
out of their control, autonomous, and competent, 
despite being able to give it tasks to perform. 
Finally, they found that (predominately female) nurses 
viewed the robot as a novelty, toy or entertainment device, 
not something that would ease their own workloads but 
instead entertain the higher status men who acquired it.

Dautenhahn et al. tested the effect of different of the direction 
by which a robot approaches a human to present a fetched item
on the participant's level of comfort~\cite{dautenhahn2006may}. 
Their studies revealed gender differences: 
a first study held outside of laboratory conditions had 
21 males and 18 females, and a second 
one held in controlled, laboratory conditions had 
9 males and 6 females. 
More females preferred a frontal approach compared 
to males, and more males preferred a right approach compared to females.
More males least preferred the front robot approach compared to females, 
who least preferred the left approach. 
However, the authors often report quantitative results without discussing 
whether they are statistically significant, and also report that some
results are significant with associated p values as large as 0.08,
(with no discussion of using an alpha level other than the
conventional 0.05)
leaving me unconvinced as to the statistical trustworthiness of this study.
At no point is there a discussion of effect size, further bringing into 
question the real world value of their results.

Skantze et al. conduct a study on participation equality
in human-robot conversations, in which 254 perceived females
and 330 perceived males interacted with a male-appearing 
anthropomorphized robot in a museum, in 
pairs~\cite{skantze2017predicting}. 
They find that interaction equality with respect
to speaking time is lower when
the two users are different with respect to age and gender, 
with the most imbalanced pairing being perceived female 
children with perceived male adults.
They find that they can also predict imbalance, 
and that interaction equality can be improved
when the robot directs questions to the user who is 
interacting less.

Halpern and Katz conducted a survey of 873 undergraduate students 
about their attitudes towards robots: one third towards a humanoid,
one third toward a doggy robot, and a final third toward an android~\cite{halpern2012unveiling}.
They find that the humanoid robots were perceived as more humanlike.
They analyzed these responses with respect to gender, religion, 
and self efficacy with technology, and find that those identifying
with Judeo-Christian religions liked robots less, 
that those with high technology self efficacy expressed more 
agreement with themes of cyber-dystopianism.
Finally, they found that self identified women like robots less,
and agree with robotophobia more than men.
They do not discuss effect sizes, do not discuss how they identify 
participant's gender, nor discuss why they assumed a religion binary
(Judeo-Christian vs not Judeo-Christian).

Marcel Heerink showed a video of an robot interacting with an old 
person of unreported gender and age to 43 female and 23 male
participants, where the robot is shown monitoring the user, 
reminding to take medication at the prescribed time, and as a fitness
advisor,
and measured participants attitudes using multiple likert scale 
questions along different axes like trust or anxiety~\cite{heerink2011exploring}. 
They found that men were more likely to have experience using computers,
and that they also have a higher perceived ease of use with the robot,
but didn't find gendered affects along other constructs. 

Shahid et al. investiate how boys and girls of 8 and 12 yeasrs of age
experience playing a collaborative game with a cat robot.
In the first study, they evaluate their change in emotional state after playing this game,
and do not investigate effects with respect to gender despite having a balanced 
girl/ boy participants~\cite{shahid2010child}.
In their second study, they evaluate whether adult men and women can 
effectively evaluate expressiveness and whether game was won from recordings of 
kids in the previous study. 
They investigate but do not find a significant main gender effect, 
but do find an interaction between age and gender:
participants find 8 year old boys easier to classify 
whether they had won or lost the game than 8 year old girls. 
Effect sizes were not discussed. 

Woods et al. studied the extent to which 14 male and 14 female 
university students preferred and felt similar, personality wise,
to a non humanoid robot which was designed to be socially 
interactive, and one which was designed to be socially 
ignorant~\cite{woods2005robot}. 
They find that participants overall did not view the robot's 
personality as similar to either style, and viewed their own 
personality characteristics as stronger. 
With respect to gender, they found that males who believed themselves
to be anxious and psychotic, the more they rated the socially
ignorant robot as also anxious and psychotic. 
They further found that females who believed themselves
to be assertive and dominant, the more they rated the socially
ignorant robot as also assertive and dominant. 
No such gender effects were noticed for the socially interactive
robot. 
The authors do not discuss effect size, bringing the real world
significance of these results into question. 
As is common in HRI studies, the authors calculate pairwise
correlations, meaning that gender effects may instead be a
correlate with a different underlying variable. 

Straight et al. conduct a study by recruiting 193 female and 317 male 
participants off of Amazon Mechanical Turk, and showing them one of
two kinds of videos: one with a robot interacting with a participant 
using polite speech, and one with a robot exhibiting direct 
speech~\cite{strait2015gender}.
They measure the extent to which participants perceived the robot
as comforting, considerate, and controlling using a multi item 
questionnaire. 
With respect to gender, they find that while using polite speech 
improved participant ratings of comfort, considerateness, and being
less controlling, this effect was stronger for female participants.
Female participants also rated the task the drawing task the robot
and human in the video were performing as less difficult, and 
expressed a greater interest in interacting with the robot. 
The authors note that their small effect sizes pose a threat to the
real world relevance of their study. 

On the dubious premise that ``The ultimate test for the life-likeness 
of a robot is to kill it'', Bartneck et al. conduct a study in which
they observed the destructive behavior on a robot of 15 male and 10
participants in which they were instructed to ``kill'' a small 
bug-like crawling robot by hitting it with a 
hammer~\cite{bartneck2007kill}. 
The study confirms that women were significantly less likely 
to break the robot into as many pieces, and to perceive the robots
as more intelligent, but with no effect on the number of times they
hit the robot. 
The authors report that women appear to have had difficulty 
handling the provided hammer. 

Otterbacher et al. showed videos of robots to 
25 male and 25 female Amazon Mechanical Turk workers, 
some of which depicted robots with male gender queues, some with
female gender queues, and some with no perceived queues~\cite{otterbacher2017s}. 
They then asked participants to rate whether they thought the robot
felt pain for fear, whether the robot had agency, 
and the participants affective response to the robot. 
They find that male and female participants 

Nomura et al. performed a study with 22 male and 31 female
university students about their negative attitudes towards robots, 
then had them talk to an anthropomorphic robot which then instructed
participants to touch it~\cite{nomura2006experimental}.
Additionally, they measure the participants distance from the robot,
how long it took them to talk to the robot initially and also 
how long it took them to respond to its question, 
and how long it took them to touch the robot after being commanded to do so.
They found that people with negative experiences is correlated with 
not talking or taking longer before talking to the robot.
They found that female participants had lower negative attitudes towards
robots having emotions (no gender effects for interacting with or the social
influence of robots), and that they initially stood
closer to the robot in the study.
The grammar of this paper made its results hard to interpret. 
The subscale on which significant results was found was negative coded,
and it is unclear if the authors properly reversed participant answers.

Mutlu et al. test the effect of robot gaze frequency on 12 male and 8 female 
participants' assesment of the robot and their recall of a story it told them, 
and found that robot gaze had a significant effect on female participants with 
respect to their recall, but males did not, and that women rated the robot 
worse when it looked at them more whereas males did the reverse~\cite{mutlu2006storytelling}. 

A 1990 experimental study by Rahimi and Karwowski found no significant 
statistical difference between male and female university participants' 
perceived robot safe speed,
but that women tended to wait longer for a robot's motion to cease
before deciding it was safe to approach it~\cite{rahimi1990human}. 

Syrdal et al. studied the effect of different approach scenarios 
on comfort for a robot approaching 20 males and 13 females for a 
one off study, and 8 males and 4 females for a 5 week longitudinal
study, measuring their comfort 3 times over 5 weeks of interaction
with the robot~\cite{syrdal2007personalized}. 
They found that men had a closer preferred approach distance when 
it is approaching from the side than the front, whereas 
no difference was observed for women.
When the robot approached from the front, females also allowed the 
robot to approach closer than males did. 
However, these effects were only witnessed on the first encounter
with the robot, and appeared to wear off as participants became more 
acclimated. 
The authors do not discuss effect size.

Oosterhout and Visser discuss a study in which they place a 
large and a small robot in a public space, and photograph 135 non 
consenting people as they interact with it as it moves,
and record the distance they interact with it at~\cite{van2008visual}. 
I didn't find any discussion of the number of female
and male participants, nor discussion of how their gender
was assigned, but I assume it was assigned from the photographs
because that is how they assigned their ages. 
They found that teens of different genders approached
the robots at significantly different distances: males approached
about 20 centimeters closer. 
They report that among adults, males approached ``significantly''
closer, despite reporting a \textit{p} 
value greater than 0.05 for this 
test, so I do not trust this study.

\subsection{Gender Of Both Robots and Participants}

Tay et al. had a robot perform healthcare and security functions, 
while manifesting different genders~\cite{tay2014stereotypes}. 
They found that their 84 male and 79 female participants tended to 
react better to the healthcare robot when it was female presenting,
and reacted better to the security robot when it was male presenting. 
This shows that robots are received better when they match the stereotype
of the task they are performing.
They report effect sizes consistently, but only discuss it to compare which 
effects are stronger. 

Otterbacher and Talias study the effect of human and robot gender with respect
to the uncanny valley (the proposition that less anthropomorphic robots are well received,
and very anthropomorphic robots are well received, but there exists some gulf in between
which is uncanny)~\cite{otterbacher2017s}.
They find that the 25 men's uncanny reactions to 25 female cueued robots 
(from YouTube) are best explained
by their perceptions of experience, whereas women's uncanny reactions to 
robots is driven by perceived agency.
They report but do not discuss effect sizes, except to compare which effects are 
stronger. 

Siegel et al. study the effect of robot and subject gender on the ability of robots to persuade
the subjects to donate money they had been given for the experiment.
They find that the 76 men were more likely to donate money to the female robot,
while the 58 women didn't show a robot gender preference~\cite{siegel2009persuasive}. 
Subjects also generally rated robots of the opposite gender as more credible, 
trustworthy, and engaging. 
While they do not discuss effect size, they report means for each group which allows
readers to evaluate this themselves.
I evaluate effect sizes to be small. 

Stafford et al. studied how people's prior robot attitudes affect their 
evaluations of a conversational robot, with 7 older men and 13 older women participants~\cite{stafford2014older}.
Among other results, they did not find that the gender of the robot affected the
interaction, but they did find that men rated robots higher in 
all conditions higher than women, with a medium effect size. 
They find that when asked to draw pictures of robots, most tended to draw male 
or gender neutral robots.

Kuchenbrandt et al. studied whether the gender typicality of a task 
would affect the extent to which 38 female and 35 male participants correctly 
performed and accepted help when performing a task when instructed by a male or 
female robot~\cite{kuchenbrandt2014keep}.
They found that participants made more errors when participants worked with the robot
to complete typically female tasks, and that after performing a typically female task,
they were less likely to accept help from the robot in future and 
anthropomorphize the robot less, showing that the gender stereotype assigned to tasks
participants perform with robots affect their perceptions and acceptance of
the robot. 
The authors present a rigorous discussion of effect sizes which stands as a shining 
example which should be emulated by other other HRI studies, noting that 
because they detect medium to large effects this underscores the practical relevance
of their findings. 

Reich-Stiebert et al. examine the influence of a robots gender teaching either a stereotypically male or female subject to either 60 male or 60 female
participants~\cite{reich2017ir}. 
They found that participants that 
robot gender does not affect participants’ learning, 
intrinsic motivation, and the evaluation of the robot.
However, they found that 
when the gender of the robot is mismatched with the 
gender typicality of the subject matter being taught, 
participants were more interested in future learning with a robot when the robots’ gender did not match the task gender typicality of a task.
The authors did not find (or did not discuss) any interaction between 
participants gender and robot gender. 

In a study by Alexander et al. 24 male and 24 female 
participants completed four Sudoku-like puzzles with a robot with a female name and voice and another with a male name and voice. 
Contrary to assumptions they form from the psychological literature,
hey found that male participants asked the robot for help more 
frequently regardless of its assigned gender. 
Participants of both genders reported 
feeling more comfortable with a robot assigned the other gender and preferred the male 
robot’s help. 
Findings indicate that gender effects can be generated in human-robot 
collaboration through relatively unobtrusive gendering methods and that they may not 
align with predictions from psychology~\cite{alexander2014asking}.

Koulouri et al. conducted an experiment in which one participant instructed
another remote participant, who they thought was a robot, on 
how to navigate to a goal using chat~\cite{koulouri2012we}. 
The second drove an on screen turtle (which the authors call a robot) 
through a map, which both participants can see.
They tested the efficiency and word use
of all combinations of male and female participants,
but in no case did the participant know the gender of the other.
The authors found that matched gender participants 
outperform mixed gender participants, 
and that males tend to employ landmark references
when interacting with females compared to 
female/female pairings or when instructing males. 

Sung et al. surveyed 379 iRobot Roomba (vacuuming robot) owners, and found that they were equally likely to be female as male, 
and use this to problematize the notion that vacuum use and 
ownership is stereotypically female whereas 
robot ownership and use is for men~\cite{sung2008housewives}.
Despite this, they found that participants were more likely 
to give Roombas and thus be more useful to women.
and further that participants who ascribed gender to their robot 
were more satisfied with the robot than those who did not, and those
who did were equally likely to refer to it as male as female, and a 
large fraction likely to ascribe both genders to it.
They also found that men and women were approximately equally likely
to name their Roombas.

Carpender et al. exposed 10 women and 9 male university students to 
videos of two robots interacting with humans, 
both introduced as designed to be a friend or member of the 
family~\cite{carpenter2009gender}. 
One was highly anthropomorphic and appeared female with skin and 
hair, and the other had eye like things and arms but looked distinctly mechanical.
They administered a three question likert scale questionarie 
about machine/humanness, friendly/unfriendly, and comfort level
of having the robot in their home, to which they found no significant results except for machine/humanness,
followed by a semi structured
interview. 
The female appearing robot's perceived gender came up 
frequently in this interview, 
with participants saying they felt more comfortable 
with it because it was female, pretty, and comforting.
Participants mainly suggested that both robots 
could do menial tasks, but felt less comfortable letting 
it do social tasks such as answering the phone 
or taking care of kids, and especially uncomfortable
with the robot touching humans in a social or affectionate way.
Unfortunately, they did not discuss how participants 
of different genders may have reacted differently, 
nor identify the gender of participants when presenting
participant quotes.

Alternately, Dautenhahn et al. conducted questionaires and 
human subject trials with 14 male and 14 female 
participants to study their perceptions of having a robot 
companion in the home~\cite{dautenhahn2005robot}.
They found that few participants wanted a robot friend,
nor did they want it to perform child or animal care tasks
instead wanting it to perform household chores. 
They wanted it to be able to communicate in a humanlike 
way, but cared less about having humanlike behavior
or appearance.
They did not find a statistical relationship between
the participant's gender and these attitudes, 
but this may be due to the small sample size,
and unfortunately while they did ask open ended
questions yielding qualitative responses, 
they did not report many results nor
analyze these results with respect to participant 
gender. 

Walters et al. conducted a human subjects study in which
a robot instructed 31 female and 37 participants in a
symposium on robotics
to approach it using either a high quality recorded male, 
female, or neutral synthesized voice, or by the 
experimenter~\cite{walters2008human}. 
They found that most participants approached 
to a zone described as ``personal'' space,
followed by a distance described as ``intimate'' space.
People approached the robot the closest in the human voice
condition (mean 42cm), and least close in the neutral synthesized 
voice (80cm), with the male and female voices separated by less
than 10 centimeters, but these effects were only
statistically significant when comparing 
the synthetic and gender neutral voice 
to all other conditions.
They also asked participants to name the robot, 
and found that the majority provided male names (41\%)
followed by neutral names (58\%), with only one percent
giving it a female name, and many refusing to give
any name at all, but they did not break out this
result with respect to the gender of the robot's voice
or the gender of the participant, nor ask why some 
refused.

Sandygulova et al. perform an observational study of children 
playing with social robots~\cite{sandygulova2016investigating}. 
They varied the gender of the voice and name 
of the social robot to 
match the reported gender of the child in some cases,
and not match it in others. 
Their study included 34 girls and 40 boys, aged 3-9 
who entered a room with a robot, 
and then performed tasks such as being greeted, 
asking how and how old the kid was, 
invited the robot to help it play in a play kitchen,
and make pretend food. 
They found that children of both genders
did not move closer or in some cases moved further away from 
the robot when it began identifying its gender by introducing
itself using a male voice and name.
When it introduced itself using a female voice and name, 
boys moved away on average.
They acknowledge that their results may be affected by the fact
that this was held in a public setting, 
and that occasionally multiple kids interacted at once,
meaning that gender matching was sometimes hard to control. 
They conclude that children recognize the gender of 
social robots, and that kids apply the same social rule of 
``gender segregated'' play to robots as they do with their 
peers. 

Rhim et al. studied the effect of changing the gender
of a robot's speech to alight with the stereotype
of the kind of task it was going to speak about in front
of participants, male for functional tasks like talking about cleaning,
and female for social tasks like encouraging the user to stretch~\cite{rhim2014effect}.
They then had this robot speak about these tasks to 15 male and 25 female
participants, and conclude that changing the gender of the voice to match the 
gender stereotype associated with the task is ``effective and efficient'', 
but do not explain how they operationalize these words, and use different words
``efficient and usefull'' when introducing their hypotheses. 
Despite their gender diverse sample, they do not investigate 
results with respect to participant gender.

Crowell et al. conduct an experiment with 23 male and 21 female
undergraduates assigned to one of four conditions: 
fully crossed design of male robot, female robot, 
male voice, and female voice (the last two with no embodiment)
~\cite{crowelly2009gendered}.
They find that male participants respond in less socially desirable 
ways in the presence of a robot than a disembodied voice,
whereas female participants behave in the reverse. 
Men and women rated the disembodied voices as more reliable
but less friendly than the embodied robot. 


The authors warn against making strong conclusions 
from their findings, but state that their results 
offer strong evidence that embodiment is important, and 
can have different effects on men and women, 
offering their past socialized experience with robots
as a explanation for why. 

Strupka et al. study the effect of a male and female voiced robot
on eight male and eight female participants on their tonal range,
volume, and other vocal measurements while answering the robots'
questions~\cite{strupka2016influence}. 
They found that both female and male participants spoke more quietly
to the female robot, even though the female robot spoke louder. 

Eyssel and Hegel study the effect of gendered facial queues (hair lengths,
lip styles of photographs of robots 
on 30 male and 30 female participants, and found that 
participants were more likely to rate the male robot as having more agency, 
and the female robot as more communal~\cite{eyssel2012s}. 
Further, they found that participants perceived stereotypically male math tasks 
as more suitable for the male robot, and verbal tasks for the female robot. 
They present this as evidence that people project traditional gendered 
norms onto gendered robots. 

Eyssel et al. test the effect of modifying the gender of a robot's voice on a 
31 women and 27 male participants' ratings of the robot's likeability, closeness, contact intentions, and 
anthropomorphism~\cite{eyssel2012activating}.
They did not find a significant effect of participant-robot gender pairing on 
likeability, but did report weak, non statistically significant evidence that 
men rated the male robot as more likeable
and that women rated the female robot as more likeable. 
They found similar evidence of varying levels of statistical significance that 
participants have more psychological closeness and contact intentions with 
robots of the same gender. 
Sadly, the authors do not discuss effect size, but by manual inspection 
different in means they report means for different genders suggest that 
there is some meaningful real world effect.

Kuo et al. evaluated how 33 female and 24 male adult participants reacted to 
a healthcare robot, named Charles and with male on-screen facial features, taking their blood pressure.
After having the robot greet the participants and having it check their
blood pressure, the participants filled out standard measures
of the quality and engagement of the social experience, and
novel measures of comfortableness~\cite{kuo2009age}. 
They found that male participants had a more positive attitude than 
females on the usefulness of the healthcare robot, 
and towards the possibility of using them in the future,
but with no significant gendered effects observed for 
social engagement or quality, or ratings of the robot. 
The authors do not discuss effect size. 

Sandygulova et al. had 56 male and 51 female children interact with 
an anthropomorphic robot 
with either a female or male synthesized voice.
They found that younger children do not successfully attribute gender
to the robot corresponding to the voice, but that older children are~\cite{sandygulova2018age}. 
Younger children indicated a preference to a robot with a matching gender,
while there was no difference in preference for a robot gender by 
older children.
The authors do not discuss sample sizes. 

Alexander et al. investigated the effect of having 24 male and 24 female 
participants complete a Sodoku puzzle while having a male or female robot that 
they can ask for help~\cite{alexander2014asking}.
They found that male participants were more likely to ask the robot
for help regardless of its assigned gender.
They also found that participants prefer to interact with a robot of the opposite
binary gender, and in general, preferred the male robot's help. 
The authors do not discuss effect size.

Powers et al. study the effect of the gender of a robot on number of words used
by 17 male and 16 female participants when describing dating norms to the robot~\cite{powers2005eliciting}. 
On the dubious assumption that females are supposed to know more about dating norms than 
males, they find that users, especially women describing norms for women, 
use more words explaining norms to a male robot than a female robot,
concluding that the assumed common ground between explainer and a female robot
leads to more efficient communication, whereas conversely, more detail is given 
to a male robot with less shared knowledge. 
The authors briefly acknowledge small effect sizes. 

Nomura and Takagi study the effect of a male and female named anthropomorphic robot
on politeness, mildness, ambitiousness and assertiveness perceived by 
17 male and 22 female students, some studying science and some studying social science~\cite{nomura2011exploring}.
They did not find main effects for robots' gender, subjects' gender, or educational 
background, but did find an interaction effect between gender and educational background:
men with science backgrounds rated the robot as more polite, with a moderate effect size.



























\balance
\bibliographystyle{ACM-Reference-Format}
\bibliography{sample-base}

\end{document}